# Detecting spills using thermal imaging, pretrained deep learning models, and a robotic platform

Gregory Yeghiyan
*Stevenson High School*
Livonia, MI, United States
gyeghiyan@yahoo.com

Jurius Azar
*Manoogian High School*
Southfield, MI, United States
jurius.azar@gmail.com

Devson Butani
*Department of Math and Computer Science*
*Lawrence Technological University*
Southfield, MI, United States
dbutani@ltu.edu

Chan-Jin Chung
*Department of Math and Computer Science*
*Lawrence Technological University*
Southfield, MI, United States
cchung@ltu.edu

*Abstract*—This paper presents a real-time spill detection system that utilizes pretrained deep learning models with RGB and thermal imaging to classify spill vs. no-spill scenarios across varied environments. Using a balanced binary dataset (4,000 images), our experiments demonstrate the advantages of thermal imaging in inference speed, accuracy, and model size. We achieve up to 100% accuracy using lightweight models like VGG19 and NasNetMobile, with thermal models performing faster and more robustly across different lighting conditions. Our system runs on consumer-grade hardware (RTX 4080) and achieves inference times as low as 44 ms with model sizes under 350 MB, highlighting its deployability in safety-critical contexts. Results from experiments with a real robot and test datasets indicate that a VGG19 model trained on thermal imaging performs best.

## I. Introduction

Spill detection is essential for public safety in dynamic environments such as cafes, restaurants, and retail spaces, where unnoticed spills frequently lead to slips and injuries. Traditional detection methods often depend on intrusive hardware or suffer from delayed response times [4], highlighting the need for fast, vision-based solutions. However, RGB-based computer vision systems are vulnerable to lighting changes and surface reflections. To address these limitations, we integrate thermal imaging and evaluate its effectiveness using pretrained convolutional neural networks (CNNs) for real-time, accurate, and lightweight spill detection. There are certain motivations for exploring the use of thermal imaging in this task.

Thermal imaging offers physical advantages that explain potential strong performance. Thermal (infrared/IR) cameras measure emitted long-wave IR radiation from surfaces rather than visible light; the recorded signal is proportional to surface temperature and emissivity. Spills produce detectable thermal signatures for several reasons.

Many spills (hot or cold beverages) have temperatures differing from the ambient floor and therefore show up as localized thermal contrasts. Because liquids generally have higher specific heat capacity (e.g. 4.186 J/g°C for water) and different thermal inertia than solid flooring materials, they either remain warmer or cooler for measurable periods after spilling, leading to visibility in thermal imaging.

Additionally, when a warm liquid contacts a cooler floor, evaporation and heat transfer create characteristic gradients at the spill boundary. These gradients increase the effective contrast at edges and make the spatial pattern of the spill more pronounced in thermal imagery than in RGB.

Liquids and common flooring materials (e.g. tile, wood) often have different emissivities as well. Even when temperature differences are small, differences in emissivity can change the intensity measured by the thermal camera, helping contrast liquid spills from the floor.

Since thermal cameras sense emitted IR rather than reflected visible light, thermal-based systems remain effective under low light, glare, or complex reflections where RGB-based detectors commonly fail; the potential impact of variable lighting conditions is decreased.

Overall, these properties help explain why thermal input often contains a cleaner, less ambiguous signal that a CNN can exploit with fewer parameters and less preprocessing.

Despite these advantages, thermal imaging is not universally superior. There are some important failure cases.

For example, if a spilled liquid has the same temperature as the surrounding floor and similar emissivity (e.g. a beverage left long enough to reach thermal equilibrium with the floor), thermal contrast vanishes and the thermal image can become indistinguishable from the background, causing missed detections. Thin films of liquid also exchange heat rapidly with the floor and may equilibrate faster than bulk spills.

Environmental heat sources may also impact results. Nearby heaters, sunlight patches, or warm machinery create thermal clutter and false positives. Reflections of thermal radiation on highly reflective surfaces and heat conduction through layered floors can produce artifacts that mimic spill-like signatures.

## II. RELATED WORK

**RGB and RGB-D Vision for Spills:** Prior computer-vision research on spill or wet-surface detection has largely focused on RGB imagery (and occasionally RGB-D data). For example, Bhutad and Patil [5] released a dataset of 1,976 labeled images of stagnant water and wet surfaces, enabling supervised learning for water/vs-dry classification. Using such RGB datasets, Gawdzik and Orłowski [9] applied Mask R-CNN to detect and segment spilled liquids in an industrial setting. Their approach trained a two-class deep detector on RGB images of known spills, achieving high accuracy in identifying liquid regions. While these methods demonstrate that CNNs can learn visual features of liquids on floors, they rely solely on appearance cues and can struggle with low contrast or reflections. Related work in robotics has used RGB-D sensors to aid water detection: Yang et al. [6] propose a polarized RGB-D (pRGB-D) framework that fuses color, polarization, and depth to detect "water hazards" (puddles) for visually impaired navigation. By combining polarization cues with depth data, the pRGB-D system can disambiguate specular reflections. Such multi-cue approaches improve robustness but add hardware complexity. In summary, pure-RGB CNN methods (e.g. Mask R-CNN) can learn to detect spills when ample labeled data is available, but have limited reliability under challenging lighting or surface conditions (e.g. dark floors, low contrast).

**Thermal Imaging for Liquids and Leaks:** Thermal infrared cameras have been studied for liquid and leak detection, leveraging the temperature contrast between liquid and surroundings. Appuhamy et al. [8] developed a thermal-camera-based method to map surface moisture. They spray liquid onto a surface and use an IR camera plus image processing to highlight areas with large temperature gradients, detecting wet coverage noninvasively. Their experiments show that liquids with $\geq 5°$C difference from the background are reliably detected. In industrial maintenance, Bao et al. [7] designed a dual-camera system combining IR and visible light to find colorless fluid leaks (e.g. water) on hot pipework. The system uses IR images to locate cooler "effusion" regions and then uses visible-light images to refine the contour of the leak. As Bao et al. explain, "the overall process first uses infrared imaging to locate a potential effusion and then conducts a secondary assessment for the final effusion characterization". This pipeline approach isolates candidate leaks via thermal contrast and then confirms them via visual processing. More broadly, thermal imaging is widely used in gas and pipeline leak detection. For example, Zhang & Zhang [11] note that thermal images (often from optical gas imaging) are "less prone to be affected by weather" and provide an orthogonal data source for leak monitoring; they feed thermal images into CNNs (AlexNet, ResNet, MobileNet, etc.) as part of a multi-sensor pipeline leak detection system. In sum, these studies show that IR cameras can reveal spills or leaks invisible in RGB (especially transparent fluids), and that combining IR with image analysis (sometimes with additional sensors) yields high detection accuracy.

**Multi-Modal Hybrid Systems:** Several works fuse RGB with other modalities to improve spill detection. As noted, Yang et al.'s pRGB-D sensor [2] integrates polarization and depth cues with RGB to detect floor water. Bao et al. [3] explicitly fuse infrared (thermal) and visible cameras: first the IR channel isolates the leak region, then the visible channel refines it. Similarly, Zhang & Zhang [7] combine thermal imaging with gas sensor data in a deep-learning architecture for pipeline leaks. These hybrid systems leverage complementary modalities: for instance, IR can distinguish cold liquid pools, while RGB provides high-resolution spatial detail. However, many existing hybrid methods use hand-engineered fusion or multi-stage pipelines. In contrast, our approach tests the efficacy of end-to-end CNNs on the combined data (e.g. stacking RGB+thermal channels) so that a single pretrained network learns from both modalities. That is, rather than separate IR processing, we feed multi-modal imagery into one model.

**Lightweight Real-Time Models:** For robot or mobile deployment, efficiency is critical. Lightweight CNN architectures (e.g. MobileNet, SqueezeNet, Tiny-YOLO) have become popular for on-device vision. For example, Bouguettaya et al. [10] survey mobile CNNs and note that MobileNet (with width=0.25, resolution=0.714) can run at ~28 FPS on an NVIDIA Jetson TX2. They also note MobileNet-SSD (an object detector) uses depthwise convolutions and achieves strong COCO accuracy, and that SSDLite (MobileNetV2 backbone) "outperforms YOLOv2 on COCO dataset with 20x more efficiency and 10x smaller" model size. Other works similarly demonstrate that reduced-channel or quantized networks can meet real-time constraints while retaining acceptable accuracy. In the context of spills, few papers explicitly focus on model size, but the insight is clear: one can deploy CNNs on robots by using such lightweight variants. Our work differs in that we use pretrained networks (like full-size ResNet or EfficientNet) for accuracy, and then evaluate their feasibility on consumer hardware and a robotic platform. By comparing performance across standard pretrained models and modalities, we assess the tradeoffs directly.

**Comparison to Our Approach:** In summary, prior methods use specialized cues or multi-stage pipelines (e.g. polarization [6], thermal+visible [7], etc.) or custom CNNs. In contrast, our approach evaluates general CNN classifiers pretrained on large datasets and fine-tuned for spill detection. We compare their performance on RGB, thermal, and fused input, explicitly measuring inference speed and accuracy. Unlike [6][7][9], which engineer modality-specific preprocessing, we investigate to what extent off-the-shelf deep models can handle spill detection under different modalities. This yields insight into whether a unified learned model can replace complex hand-tuned pipelines.

## III. DATASET & EXPERIMENTAL SETUP

To be able to create a model to detect spills, it is first necessary to collect sample images to train the model on. Since standard RGB cameras struggle to distinguish the floor from water spills, similar to human vision, data was collected with a combination of a thermal camera and an RGB camera. A Topdon TC001 was used as the thermal camera, while a Genius

WideCam F100 was used as the RGB camera. A Python script was used to save images from the thermal camera and RGB camera at the same time, which can be found in the following GitHub repository: https://github.com/Aeolus96/ThermalCameraCapture. Spill images were put into the spill folder and images without spills were put into the no-spill folder. By the end of the research, 4000 total images were collected, evenly split across the image types RGB and thermal, and evenly split across the classes spill and no-spill. Four liquids - water, Coke, red juice, and yellow juice - were used, and two rooms were used in data collection: Atrium and J234. This resulted in 8 combinations (Room x Liquid x Modality). A variety of spill sizes were collected, forming a diverse spill dataset. Within these sizes, a typical small spill would have a diameter of 2-4 inches, with the regions being approximately circular. A typical large spill would have a diameter of up to 12 inches assuming an approximately circular region. Over time, these large spill regions deformed as the liquid flowed across the floor, which led to an increase in certain dimensions. The atrium had porcelain tiled floors, while J234 had polished concrete flooring. Foot traffic was not included in pictures in the dataset. In the atrium, lighting conditions were kept consistent, while lighting conditions in J234 were dynamic due to the high presence of natural lighting and variability of sunlight.

Data collection was performed using a robot with a thermal and RGB camera mounted to it. A Lenovo Legion Pro 7i with NVIDIA RTX 4080 was used for control and image collection. The cameras were plugged directly into the laptop, and the robot was controlled using a USB joystick plugged in to the laptop with inputs being relayed to the robot through serial communication. Images were collected from a variety of positions and angles, and in isolation from environmental heat sources. Robotic inspection using thermal vision has proven effective in industrial environments for leak detection and has been implemented in pipeline inspection systems [1].

To split the data, another script was used on collected images. The dataset was split 70-20-10 (training, validation, test). Another script was used to match the view of RGB images and thermal images through cropping and a perspective transformation. Additionally, for the purpose of experimentation, a third modality was introduced: side-by-side combined images. Transformed RGB images and thermal images were combined through side-by-side concatenation, with thermal on the left and RGB on the right. The resolutions were 256x192 for thermal, 640x360 for RGB, and 512x192 for side-by-side (RGB was downscaled to 256x192 prior to concatenation).

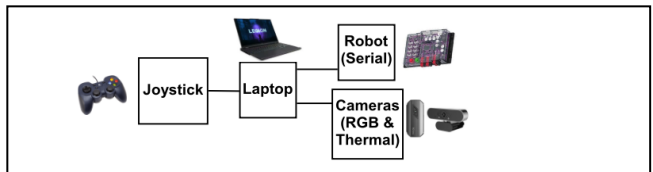

Fig. 1. A diagram of the data collection and testing setup

## IV. METHODOLOGY AND EXPERIMENTATION

**Training:** In order to determine whether RGB, thermal, or combined images would yield the best accuracy and efficiency, a VGG19 model was trained on each type, and the resulting model was tested on a test set and real-time input. The same hardware was used to train the model as was used to collect data: A Lenovo Legion Pro 7i with NVIDIA RTX 4080. Data used and real-time testing area were homogeneous across room and liquid types for this comparison

For the training strategy, the last 5 layers were fine-tuned, an RMSprop optimizer was used (lr=1e-5), a binary crossentropy loss function was used, and early stopping was used with patience = 5. A batch size of 8 was used for training and validation, while a batch size of 2 was used for test. The model was trained with up to 50 epochs; however, this was never reached due to the application of the early stopping mechanism. For data augmentation, images were randomly flipped horizontally, slightly rotated, and had small variations in contrast during training, The random rotation and contrast had factor 0.01. The standard VGG preprocessing pipeline was then used. The results are found in Table I.

**Testing:** The results show that Combined, although accurate on the test dataset, is not effective in real-time testing. RGB, although effective in real-time testing and near-perfect in the test set, is slower for inferences and has a model size three times that of thermal. Thermal performs with 100% accuracy on the test set and real-time testing, along with offering the shortest inference times and the smallest model size. Our results align with recent findings that show the effectiveness of deep convolutional architectures in leak detection tasks using thermal imaging [2].

We proceed by probing the efficacy of the thermal-VGG19 configuration by creating individual models for each room-liquid combination and noting accuracy on test datasets and real-time testing. The same training strategy was employed. The results are found in Table II.

Based on the high accuracy consistently indicated throughout these results, we note that a thermal-VGG19 configuration has high efficacy in a variety of situations, given isolation from environmental heat sources. This is likely due to the homogeneity of no-spill images when environmental heat sources are not introduced. Differentiation between the two classes may potentially be reduced to identifying changes in the homogeneity of thermal input.

**Training:** In order to identify pretrained models with higher time efficiency without sacrificing accuracy, a collection of various models were trained on the comprehensive data set, and their accuracy over the test data set was logged. The results can be found in the first two columns of Table III.

**Real-time Testing:** Although all model choices performed with high accuracy, Nasnet Mobile seems promising, seeing that it was the only model that could match VGG19's 100% accuracy. In order to determine which model would be better for a real life scenario, real-time testing was performed to compare inference times, with the goal of selecting the most efficient and lightweight model. The results can be found in the last three columns of Table III.

TABLE I. RGB VS THERMAL VS COMBINED

| Image Type | % Test | Demo Accuracy | Model Size | Inference Time |
|---|---|---|---|---|
| *Thermal* | 100% | 100% | 324.6 MB | 44 ms |
| *RGB* | 98.84% | 100% | 1.0 GB | 55 ms |
| *Combined* | 100% | 60% | 525.9 MB | 47 ms |

TABLE II. THERMAL-VGG19 ACCURACY

| Room | Liquid | % Test | Demo Accuracy | Model Size | Inference Time |
|---|---|---|---|---|---|
| *Atrium* | Water | 100% | 100% | 324.6 MB | 44 ms |
| *Atrium* | Coke | 100% | 100% | 324.6 MB | 45 ms |
| *Atrium* | Red Juice | 100% | 100% | 324.6 MB | 45 ms |
| *Atrium* | Yellow Juice | 100% | 100% | 324.6 MB | 45 ms |
| *J234* | Water | 100% | 100% | 324.6 MB | 45 ms |
| *J234* | Coke | 100% | 100% | 324.6 MB | 45 ms |
| *J234* | Red Juice | 100% | 100% | 324.6 MB | 45 ms |
| *J234* | Yellow Juice | 100% | 100% | 324.6 MB | 45 ms |
| *Both* | All | 100% | 100% Atrium, 100% J234 | 324.6 MB | 44 ms |

Demo accuracy, model size, and inference time are omitted for models other than VGG19 and Nasnet Mobile in Table III, since a comparison is made only between VGG19 and Nasnet Mobile due to accuracy on the test dataset being lower for other models.

TABLE III. MODEL PERFORMANCE

| Model | % Test | Demo Accuracy | Model Size | Inference Time |
|---|---|---|---|---|
| *VGG19* | 100% | 100% | 324.6 MB | 46 ms |
| *ResNet50* | 99.66% | - | - | - |
| *ResNet50V2* | 99.22% | - | - | - |
| *EfficientNetB3* | 99.15% | - | - | - |
| *InceptionV3* | 98.88% | - | - | - |
| *Inception Resnet V2* | 99.30% | - | - | - |
| *Xception* | 99.64% | - | - | - |
| *Densenet121* | 99.66% | - | - | - |
| *Nasnet_Mobile* | 100% | 100% | 440.3 MB | 55 ms |
| *EfficientNetV2B3* | 99.85% | - | - | - |
| *ConvNext Base* | 99.85% | - | - | - |

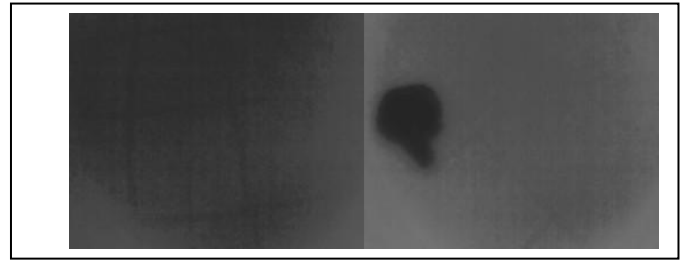

Fig. 2. Side-by-side comparison of no-spill and spill thermal images

It is found that a model using Nasnet Mobile yields greater inference times by 9 ms on average in the trials, and a greater model size by 115.7 MB as well. VGG19 performs with quicker inferences and smaller model size while achieving the same accuracy. It is determined that VGG19 remains a more appropriate model than Nasnet Mobile for this application.

V. RESULTS AND CONCLUSION

Through our experimentation, it is found that, in environments with diverse lighting conditions and isolation from environmental heat sources, a VGG19 image classification model trained on thermal imaging offers the best performance, as measured by inference time, test accuracy, and accuracy in real-time deployment.

VI. FUTURE WORK

In future work, it may be useful to perform experiments without isolation from environmental heat sources such as foot traffic. Future directions include exploring ensemble methods that fuse RGB and thermal features, as motivated by recent surveys on multi-sensor data fusion in leak detection systems [3]. This could potentially allow for misclassifications of environmental heat sources on thermal imaging to be corrected through distinction in RGB imaging.

## APPENDIX

Below is a photograph of the robotic setup. Following that are additional random sample images from the raw, unscaled dataset. They are separated by modality, room, image type, and liquid.

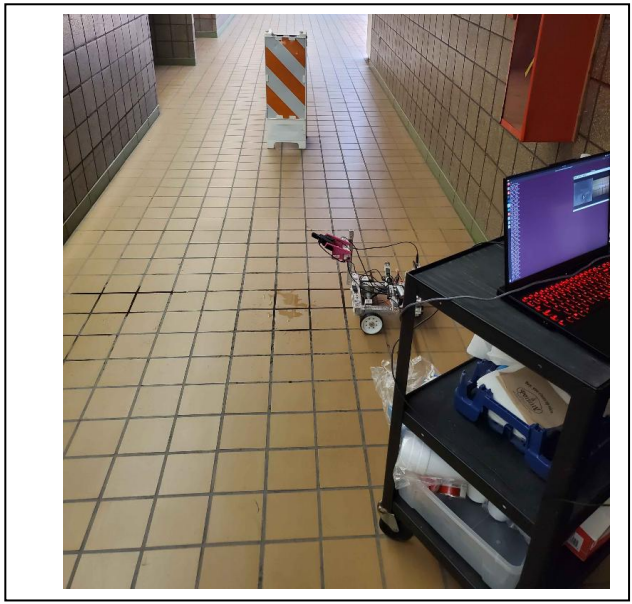

SAMPLE IMAGES

| Atrium | |
|---|---|
| **No-Spill** | **Spill** |
| Coke, RGB | Coke, RGB |
| Coke, Thermal | Coke, Thermal |
| Red Juice, RGB | Red Juice, RGB |
| Red Juice, Thermal | Red Juice, Thermal |
| Water, RGB | Water, RGB |
| Water, Thermal | Water, Thermal |
| Yellow Juice, RGB | Yellow Juice, RGB |
| Yellow Juice, Thermal | Yellow Juice, Thermal |

| J234 | |
|---|---|
| No-Spill | Spill |
| 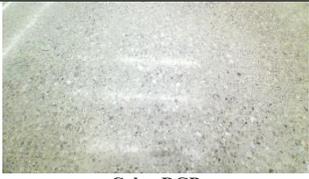 Coke, RGB | 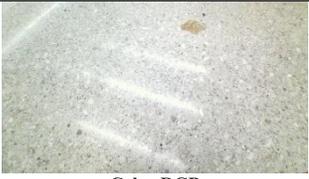 Coke, RGB |
| 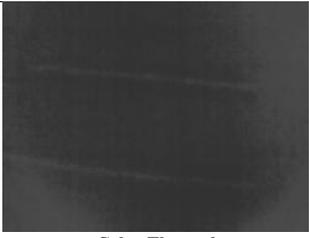 Coke, Thermal | 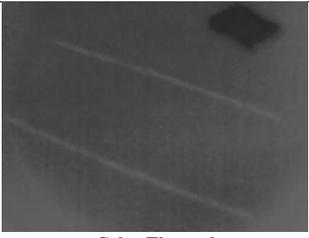 Coke, Thermal |
| 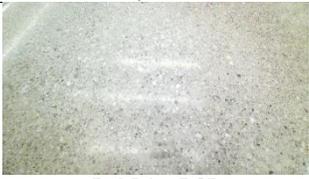 Red Juice, RGB | 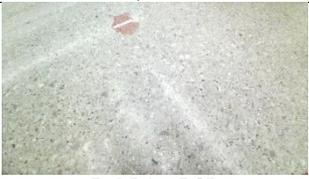 Red Juice, RGB |
| 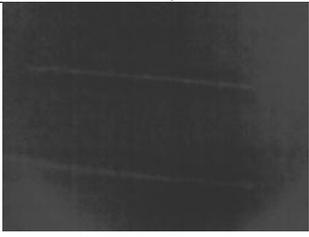 Red Juice, Thermal | 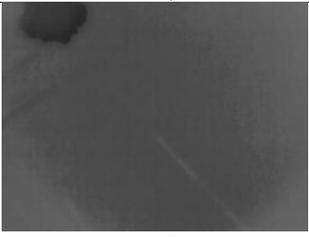 Red Juice, Thermal |
| 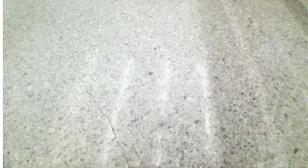 Water, RGB | 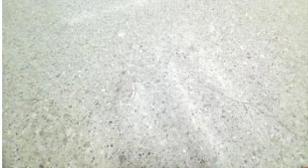 Water, RGB |
| 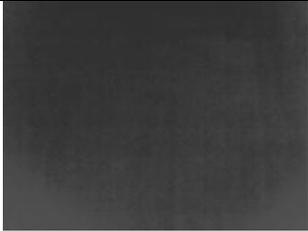 Water, Thermal | 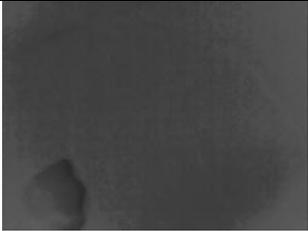 Water, Thermal |
| 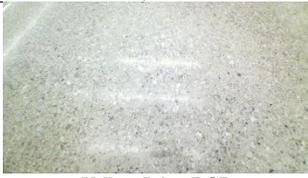 Yellow Juice, RGB | 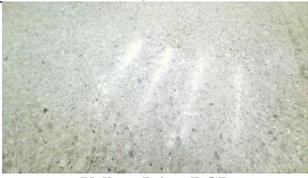 Yellow Juice, RGB |
| 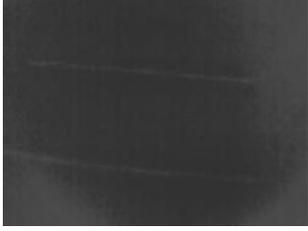 Yellow Juice, Thermal | 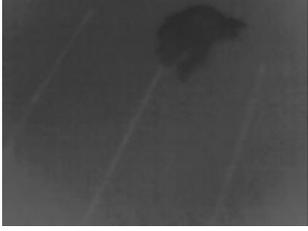 Yellow Juice, Thermal |